\pdfoutput=1

\newif\ifICML
\ICMLtrue  

\ifICML

    \documentclass{article}

\usepackage{microtype}
\usepackage{graphicx}
\usepackage{subfigure}
\usepackage{booktabs} 
\usepackage{hyperref}
\usepackage{url}
\usepackage{colortbl}
\usepackage{nicefrac}       %
\usepackage{microtype}      %
\usepackage[dvipsnames]{xcolor}         %
\usepackage{multirow}
\usepackage{multicol}
\usepackage{graphicx}
\usepackage{xspace}
\usepackage{amsmath}
\usepackage{adjustbox}
\usepackage{tcolorbox}
\usepackage{amssymb}
\usepackage{enumitem}
\usepackage{wrapfig}
\usepackage{xcolor}
\usepackage{subcaption}    %
\usepackage{float}
\usepackage{stfloats}
\usepackage{amsmath}
\usepackage{listings}
\usepackage{mdframed}
\usepackage{tcolorbox}
\usepackage{arydshln}
\usepackage{booktabs} 
\usepackage{color,soul}
\usepackage{makecell}
\usepackage[numbers]{natbib}
\usepackage{caption}
\usepackage{tabularx}
\usepackage{tcolorbox}
\tcbuselibrary{breakable}
\usepackage{hyperref}
\usepackage{url}
\usepackage{hyperref}

\usepackage[accepted]{icml2025}
\usepackage{amsmath}
\usepackage{amssymb}
\usepackage{mathtools}
\usepackage{amsthm}

\usepackage[capitalize,noabbrev]{cleveref}

\theoremstyle{plain}

\theoremstyle{definition}

\theoremstyle{remark}

\usepackage[textsize=tiny]{todonotes}
\definecolor{NvidiaGreen}{RGB}{118, 185, 0}

\icmltitlerunning{Submission and Formatting Instructions for ICML 2025}
\begin{document}
\twocolumn[
\icmltitle{The Challenge of Teaching  Reasoning to LLMs Without RL or Distillation}





\begin{icmlauthorlist}
\icmlauthor{Wei Du}{nv}
\icmlauthor{Branislav Kisa\v{c}anin}{nv,ivi,ftn,awe}
\icmlauthor{George Armstrong}{nv}
\icmlauthor{Shubham Toshniwal}{nv}
\icmlauthor{Ivan Moshkov}{nv}
\icmlauthor{Alexan Ayrapetyan}{nv}
\icmlauthor{Sadegh Mahdavi}{nv}
\icmlauthor{Dan Zhao}{nv}
\icmlauthor{Shizhe Diao}{nv}
\icmlauthor{Dragan Ma\v{s}ulovi\'{c}}{dmi}
\icmlauthor{Marius Stanean}{awe}
\icmlauthor{Advaith Avadhanam}{mit}
\icmlauthor{Max Wang}{upenn}
\icmlauthor{Ashmit Dutta}{uiuc}
\icmlauthor{Shitij Govil}{gatech}
\icmlauthor{Sri Yanamandara}{uiuc}
\icmlauthor{Mihir Tandon}{uiuc}
\icmlauthor{Sriram Ananthakrishnan}{uchi}
\icmlauthor{Vedant Rathi}{uiuc}
\icmlauthor{David Zhang}{uiuc}
\icmlauthor{Joonseok Kang}{uiuc}
\icmlauthor{Leon Luo}{uchi}
\icmlauthor{Titu Andreescu}{awe}
\icmlauthor{Boris Ginsburg}{nv}
\icmlauthor{Igor Gitman}{nv}
\end{icmlauthorlist}


\icmlaffiliation{nv}{Nvidia Corporation}
\icmlaffiliation{ivi}{Institute for Artificial Intelligence Research and Development of Serbia}
\icmlaffiliation{ftn}{Faculty of Technical Sciences, University of Novi Sad}
\icmlaffiliation{awe}{AwesomeMath}
\icmlaffiliation{dmi}{Department of Mathematics and Informatics, Faculty of Sciences, University of Novi Sad} 
\icmlaffiliation{uiuc}{University of Illinois Urbana-Champaign}
\icmlaffiliation{uchi}{University of Chicago}
\icmlaffiliation{gatech}{Georgia Institute of Technology}
\icmlaffiliation{upenn}{Department of Computer and Information Science, School of Engineering and Applied Science, University of Pennsylvania}
\icmlaffiliation{mit}{Massachusetts Institute of Technology}
\icmlcorrespondingauthor{Wei Du}{wedu@nvidia.com}

\icmlkeywords{Machine Learning, ICML}

\vskip 0.3in
]



\printAffiliationsAndNotice{}  


\begin{abstract}

Reasoning-capable language models achieve strong performance in complex tasks by generating explicit and long Chain-of-Thought (CoT) traces. While prior work has shown that such reasoning can be learned through reinforcement learning or distillation from stronger models, we ask whether long CoT reasoning can be induced in base models with only prompting or minimal supervision. We demonstrate that as few as 20 high-quality CoT examples from the reasoning model \texttt{QwQ-32B-Preview} are sufficient to transform the base model \texttt{Qwen2.5-32B} into a reasoning model via fine-tuning. This result suggests that even extremely small-scale, high-quality reasoning supervision can unlock strong generalization capabilities. We further explore CoT supervision from non-reasoning models and human annotators, enhanced through prompt engineering, multi-pass editing, and structural guidance. However, these alternatives still fall short of the CoT traces from the reasoning model, suggesting that certain latent qualities in model-generated reasoning are not easily replicated. We also analyze key data attributes, such as problem difficulty, solution diversity, and output length, that contribute to effective reasoning transfer. To facilitate future research, we release our human-authored dataset across all refinement stages.

\end{abstract}

\section{Introduction}
Recent advances in large language models (LLMs) have led to remarkable progress in natural language understanding, code generation, and problem solving. Among the most notable developments is the emergence of reasoning models, such as  OpenAI o1 \cite{openai2024o1}, DeepSeek R1 \cite{guo2025deepseek}, and  Qwen QwQ \cite{qwq32b}, which demonstrate strong performance on complex tasks by producing long Chain-of-Thought (CoT) reasoning traces. Unlike traditional LLMs, which often produce direct or short CoT \cite{wei2022chain} responses, reasoning models can engage in exploratory reasoning processes. These traces typically include stages such as hypothesis generation, intermediate verification, and self-reflection behaviors that are widely regarded as key to their superior performance on complex tasks like advanced mathematics and competition-level programming.

Existing literature often associates reasoning models with the generation of long CoT, which are assumed to reflect strong problem-solving or “thinking” capabilities. However, early work \cite{wei2022chain} showed that short CoT prompting without any fine-tuning, can improve the reasoning abilities of LLMs.
This observation motivates a central question: Can we induce reasoning mode characterized by reflective and self-exploration patterns in a non-reasoning base model using only a handful of high-quality long CoT examples?  Specifically, we ask whether a small number of such examples (e.g., 10 to 50), provided via prompting or supervised fine-tuning (SFT), is sufficient to shift the model toward a reasoning mode characterized by stronger self-exploration and reflective problem solving.

To experimentally evaluate whether a base model is induced to reasoning mode, we first introduce a concrete and operational definition of reasoning behavior in the context of mathematical problem solving: A base model is said to exhibit reasoning behavior, or equivalently to operate in a reasoning mode, if after minimal supervision under long CoT reasoning patterns, it significantly outperforms a much larger non-reasoning model on challenging math tasks. Since such a small number of examples is unlikely to convey substantial new factual knowledge, we interpret any observed performance gains not as a result of memorization or scale, but rather as evidence of a qualitative shift in problem-solving strategy, toward more iterative, exploratory, and reflective reasoning.

To test this hypothesis, we begin by evaluating the effectiveness of reasoning data. We use \texttt{QwQ-32B-Preview} \cite{qwq32b} to generate high-quality CoT traces and fine-tune the base model \texttt{Qwen2.5-32B} \cite{qwen2025qwen25technicalreport}. To evaluate the reasoning capabilities of the resulting checkpoints, we adopt the Comp-Math-24-25 Benchmark \cite{moshkov2025aimo}, a challenging dataset consisting of 256 mathematical problems requiring single numerical answers. These problems are drawn from the AIME and HMMT competitions held in 2024 and 2025. Surprisingly, we find that as few as 20 long CoT examples are sufficient to activate reasoning behavior in the base model. After fine-tuning on just these examples, the model achieves a 5.38\% gain in \textbf{pass@1} and a 11.59\% gain in \textbf{maj@64} over \texttt{Qwen2.5-Math-72B-Instruct} \cite{yang2024qwen25mathtechnicalreportmathematical}, one of the strongest open-source non-reasoning models.

Motivated by this result, we next explore whether similar reasoning behavior can be induced using solutions generated by non-reasoning models. Compared to human-written reasoning data, which are expensive and time-consuming to produce, non-reasoning data can be generated at scale with minimal effort. In this study, we select \texttt{Qwen2.5-32B-Instruct} \cite{qwen2025qwen25technicalreport} as a representative non-reasoning model, and explore various prompt formats and post-editing strategies using LLMs to generate long CoT solutions. Despite extensive efforts, we find it extremely challenging to generate solutions that exhibit genuinely reflective reasoning patterns. Even when using LLMs to post-edit non-reasoning outputs by explicitly inserting steps such as self-validation or reflection, the resulting traces often remain shallow and lack meaningful iterative reasoning. SFT on such data fails to activate reasoning behavior in the base model \texttt{Qwen2.5-32B} . Despite using orders of magnitude more data than in the distilled reasoning data, the resulting model only matches the performance of \texttt{Qwen2.5-Math-72B-Instruct}, and shows no signs of reasoning behavior.

Since non-reasoning data fail to induce reasoning behavior, we next investigate whether carefully designed human-written solutions can serve as an effective alternative. For this purpose we asked a group of volunteers, all of them former competitive math problem solvers, now college students, professors, and engineers, to solve a set of 50 problems. We documented four iterations of this effort - four versions of the dataset we call {\bf Nemotron-Math-HumanReasoning} - as the solutions evolved over time. We provide more details about this dataset and its versions in Sections \ref{sec:problem_sets} and \ref{sec:human_written_data} and open-source the data so other researchers can use this dataset as their starting point. Although we conducted several rounds of iterative refinement, SFT on human-written data still failed to shift the base model into reasoning mode. 

To better understand what makes reasoning data effective, we performed a series of ablation studies focusing on various factors, including problem difficulty and diversity, solution length and correctness, and the presence of reasoning-related keywords. Interestingly, across all ablation settings, we observe that the reasoning data consistently succeed in activating reasoning behavior in the base model. We hypothesize that the underlying structure and demonstration patterns of the reasoning traces remain largely consistent across these variations, which may be the key factor responsible for triggering reasoning capabilities. Although we have found that as few as 20 long CoT reasoning examples with very limited SFT, can activate the model’s reasoning mode, we were unable to replicate this behavior using non-reasoning or human-written data, even after multiple rounds of refinement. Nevertheless, we remain optimistic that carefully curated human-written reasoning data, even in small quantities, could achieve similar effects. To support future research, we release our human-labeled dataset across multiple refinement stages and encourage the community to explore further what truly enables reasoning to emerge in LLMs.

\section{Reasoning with Minimal Supervision}
\label{sec:data}
\subsection{Training details}
We perform SFT on \texttt{Qwen2.5-32B}, following a consistent training pipeline across all experiments. We conducted a grid search over various hyperparameter settings (Appendix \ref{sec:grid_search}) and adopted the configuration that yielded the best performance. We employ the AdamW optimizer~\cite{loshchilov2017decoupled} with a fixed learning rate of 1e-5 and no warmup steps. Training is conducted with a batch size of 1,024 samples, and we adopt sequence packing and context parallelization techniques from NeMo-Aligner~\cite{shen2024nemo} to accelerate learning on long-context reasoning data. For datasets with fewer than 1,024 samples, the entire dataset is used as a batch. The context parallel size is set to 4, the tensor model parallel size is set to 8, and the packing length is fixed at 16,384 tokens. Each model is fine-tuned for 50 steps, and we use the final checkpoint as the output model. All of the experiments are done using NeMo-Skills\footnote{\url{https://github.com/NVIDIA/NeMo-Skills}}.

\subsection{Evaluation Setup} 
\label{sec:comp-math-24-25}

To evaluate the performance of the reasoning models, it is essential to select a challenging dataset. We adopt the Comp-Math-24-25 Benchmark \cite{moshkov2025aimo}, which comprises 256 advanced-level mathematical problems that require single numerical answers, making evaluation both simpler and more accurate. As a non-reasoning baseline, we evaluate \texttt{Qwen2.5-Math-72B-Instruct}, one of the strongest open-source math models currently available. We generate solutions using 64 random seeds, with a temperature of 0.7, a top-p value of 0.95, and a maximum generation length of 16,384 tokens, even though, for this model, the longest correct solution observed on this benchmark is only 8,910 tokens. Evaluation is based on two metrics: \textbf{pass@1}, which measures the average accuracy across 64 independent runs, and \textbf{maj@64}, which reflects the accuracy of the majority vote among the 64 generations. The results for \texttt{Qwen2.5-Math-72B-Instruct} are shown in Figure~\ref{fig:performance_varying_number} as the dashed blue line (\textbf{pass@1}) and the dashed orange line (\textbf{maj@64}). Throughout this paper, all models are evaluated using the same configuration.

\subsection{Seed Problem Set for our Dataset}
\label{sec:problem_sets}

The {\bf Nemotron-Math-HumanReasoning}\footnote{\url{https://huggingface.co/datasets/nvidia/Nemotron-Math-HumanReasoning}} dataset begins with 50 randomly selected math problems sourced from the AoPS forums \cite{AoPS_Forums}. These problems vary in difficulty but generally fall within the range of AIME to USAMO levels \cite{AMC_MAA}. Most of them align with standard high school competition topics, such as number theory, combinatorics, algebra, and geometry, though a few also involve functions and calculus. All problems demand substantial reasoning and multi-step derivations to arrive at the final answer. Below is a representative example from the dataset:

\noindent 
{\em Find all integral solutions of the equation $x^{2}\!+\!2y^{2}\!=\!z^{2}$.}

{
\noindent 
The final answer after the full derivation is:
\[ (x, y, z) = (\lvert 2a^2 - b^2 \rvert k, 2abk, (2a^2 + b^2) k) \] where $b$ is odd, $\gcd(a, b) = 1$, and $k \in \mathbb{Z}$.

\subsection{Inducing Reasoning with Minimal Supervision}
\label{sec:reasoning_data_generation}
It is worth exploring whether a base model can exhibit reasoning capabilities when exposed to only a small amount of high-quality supervision. To investigate this, we conducted experiments using reasoning data generated by \texttt{QwQ-32B-Preview}, varying the number of training examples from 10 to 50. Specifically, we used the same fixed set of 50 math problems as described in Section~\ref{sec:problem_sets}, and generated corresponding solutions using \texttt{QwQ-32B-Preview}. For each problem, we generated 512 solutions using different random seeds and selected a single solution with the correct final answer. We thus obtained one long CoT reasoning solution per problem, resulting in 50 solutions in total. From this set, we randomly sampled 10, 20, 30, 40, and 50 examples to fine-tune the base model \texttt{Qwen2.5-32B}. 

To ensure a fair comparison with human-written and non-reasoning data in subsequent experiments, we sampled or filtered the data to select solutions whose lengths are approximately close to 3K tokens. Although not exactly 3K, the average token lengths across all datasets with varying numbers of examples range from 3,400 to 3,800. This setup enables us to assess whether a small number of high-quality reasoning demonstrations can effectively elicit reasoning behavior in the base model. As illustrated in Figure \ref{fig:performance_varying_number} (solid blue line for \textbf{pass@1} and solid orange line for \textbf{maj@64}), the model notably outperforms the baseline \texttt{Qwen2.5-Math-72B-Instruct} with just 20 training examples, achieving a 5.38\% absolute gain in \textbf{pass@1} (increasing from 11.72\% to 17.10\%) and an 11.59\% improvement in \textbf{maj@64} (from 16.14\% to 27.73\%). 
This observation aligns with findings from~\cite{muennighoff2025s1},  which report that a limited amount of high-quality reasoning data, e.g., a dataset with just 1K examples, can substantially improve performance on mathematical and coding reasoning tasks. We also repeated the SFT experiments on \texttt{Qwen2.5-7B} and \texttt{Qwen2.5-14B} using the same data and settings. However, neither model is able to effectively enter reasoning mode, with SFT performance comparable to or worse than the baseline (see Appendix~\ref{sec:sft_7b_14b}). Therefore, we use \texttt{Qwen2.5-32B} for all experiments presented in this paper.

\section{Generating Reasoning Data from Non-Reasoning Models}
Motivated by this result, we first investigate whether similar reasoning behavior can be induced using solutions generated by non-reasoning models, as such data is easy to generate at scale and incurs minimal cost. We use \texttt{Qwen2.5-32B-Instruct} ~\cite{qwen2025qwen25technicalreport} as the non-reasoning model to collect long CoT generations. We selected \texttt{Qwen2.5-32B-Instruct} because it follows instructions more effectively than math-specific large language models such as \texttt{Qwen2.5-Math-72B-Instruct}.

\subsection{Prompting Strategies for Generating Synthetic Reasoning Data}
To better elicit long CoT outputs from the non-reasoning model \texttt{Qwen2.5-32B-Instruct}, we designed two prompting strategies (see Appendix \ref{sec:data_generation_instruction} and \ref{sec:data_generation_few_shot}): one based solely on detailed instructional guidance, and another that augments the same instructions with the same 20 explicit reasoning examples as Section \ref{sec:reasoning_data_generation} used to conduct SFT on \texttt{Qwen2.5-32B}. 
Using the 50 problems described in Section~\ref{sec:problem_sets}, we generated 1,024 responses per problem using different random seeds. We retained only those outputs that produced the correct final answer and were between 1K and 8K tokens in length. This filtering yielded 2,499 valid samples from the instruction-only prompt (Appendix \ref{sec:data_generation_instruction}) and 4,953 from the example-augmented prompt (Appendix \ref{sec:data_generation_few_shot}). According to the first two rows of Table~\ref{tab:eval-results-non-reasoning-data}, the example-augmented prompt generates a greater amount of data, with an average length of 1,214, slightly exceeding the 1,162 average from the instructional-only prompt.

We then fine-tuned the base model, \texttt{Qwen2.5-32B}, on each of these datasets. The performance of the resulting models on Comp-Math-24-25 is shown in the first two rows of Table~\ref{tab:eval-results-non-reasoning-data}. Although we used significantly more than 50 solutions, both prompts resulted in similar performance from the baseline \texttt{Qwen2.5-Math-72B-Instruct}. This suggests that prompting a non-reasoning model to generate sufficiently high-quality training data remains challenging and may be insufficient to transition the base model into a reasoning-capable regime. For comparison, the last two rows of Table~\ref{tab:eval-results-non-reasoning-data} show the results of in-context learning using the same 20 examples as few-shot prompts for \texttt{Qwen2.5-32B} and \texttt{Qwen2.5-32B-Instruct}, which perform worse than the SFT counterparts.

\begin{figure}[t]
    \centering
        \includegraphics[width=0.5\textwidth]{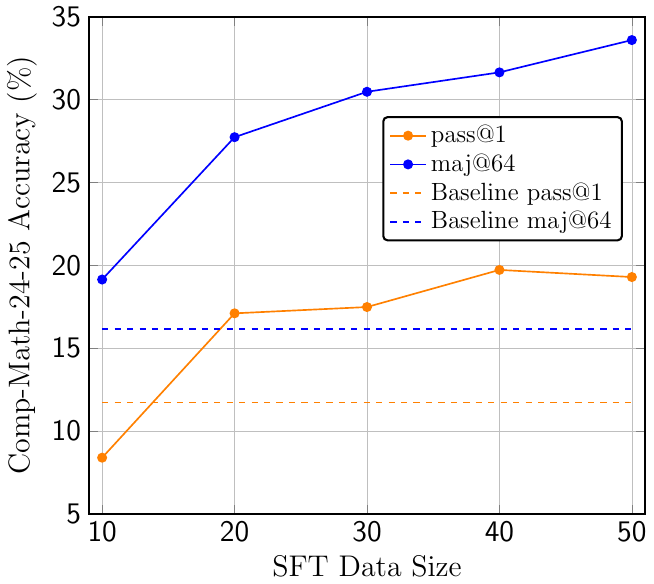}
        \label{fig_cot_genrm}
    \caption{SFT accuracy of \texttt{Qwen2.5-32B} on Comp-Math-24-25 under varying data sizes. With only 20 examples, it outperforms the baseline \texttt{Qwen2.5-Math-72B-Instruct}. All datasets have average token lengths between 3,400 and 3,800. }
    \label{fig:performance_varying_number}
\end{figure}

\subsection{Editing Synthetic Solutions to Induce Reasoning Patterns}
\label{sec:solution_edit_non_reasoning_data}
The previous section demonstrates that prompting alone is insufficient to induce non-reasoning models to generate solutions exhibiting structured reasoning, such as self-verification, intermediate checks, or illustrative examples. Consequently, the resulting training data lack the necessary reasoning characteristics to transform the base model into reasoning model. This limitation raises a natural question: can LLMs be leveraged to revise existing non-reasoning solutions, enriching them with reasoning characteristics such as self-validation, self-check, or reflective examples?

To explore this, we used \texttt{Qwen2.5-32B-Instruct} to edit the original solutions via prompt-based instruction. We designed three prompts to encourage the insertion of reasoning behaviors: (i) one containing only detailed instructional guidance (Appendix~\ref{sec:data_edit_instruction}); (ii) one combining the same guidance with the 20 explicit reasoning examples used in Section \ref{sec:reasoning_data_generation} (Appendix \ref{sec:data_edit_few_shot}); and (iii) one instructing the model to incorporate an incorrect attempt into a revised reasoning-based solution (Appendix~\ref{sec:data_edit_combination}). For the third prompt, we randomly selected an incorrect response from the same data pool and paired it with a correct solution, prompting the LLM to merge them into a single, improved response that reflects both reasoning and correction.

For data editing, we used the solutions generated by the example-augmented prompt (Appendix~\ref{sec:data_generation_few_shot}). Rows 3 - 5 of Table~\ref{tab:eval-results-non-reasoning-data} report the SFT performance on these revised solutions. Although the average length of the revised solutions increased compared to the original ones, indicating the insertion of some reasoning patterns, the resulting fine-tuned models showed no meaningful improvement. Their performance remained comparable to both the baseline model and the model fine-tuned on the original solutions. This suggests that the inserted reasoning patterns were superficial and insufficient to activate the base model’s reasoning capabilities.


\begin{table*}[h]
\centering
\renewcommand{\arraystretch}{1.3} 
\begin{tabular}{rlcccc}
\hline
   & \textbf{Data} & \textbf{Pass@1} & \textbf{Maj@64} & \textbf{Size} & \textbf{Average Length} \\

\hline
 1 & Non-reasoning data (Prompt \ref{sec:data_generation_instruction})  & 12.26\%  & 14.84\%  & 2,499 &  1,162   \\
 2 & Non-reasoning data (Prompt \ref{sec:data_generation_few_shot})     & 11.49\%  & 14.45\%  & 4,953 &  1,214   \\
 3 & Non-reasoning data edit (Prompt \ref{sec:data_edit_instruction})   & 12.08\%  & 15.62\%  & 4,953 &  1,384   \\
 4 & Non-reasoning data edit (Prompt \ref{sec:data_edit_few_shot})      & 12.36\%  & 15.75\%  & 4,953 &  1,345   \\
 5 & Non-reasoning data edit (Prompt \ref{sec:data_edit_combination})   & 11.10\%  & 15.23\%  & 4,953 &  1,406   \\
\hline

 6 & Human-written data ver.\,1                                                     & 5.51\% & 15.23\% & 50 &   2,679 \\ 
 7 & Human-written data ver.\,2                                                     & 5.04\% & 13.28\% & 50 &   2,840 \\ 
 8 & Human-written data ver.\,3                                                     & 3.88\% & 10.94\% & 50 &   3,088 \\ 
 9 & Human-written data ver.\,4                                                     & 4.47\% & 13.28\% & 50 &   3,171 \\
10 & Human-written data ver.\,4 edit (Prompt \ref{sec:data_edit_instruction}) & 7.82\%  & 17.97\%      & 50 &   2,591 \\
11 & Human-written data ver.\,4 edit (Prompt \ref{sec:data_edit_few_shot})    & 10.03\%     & 15.62\%      & 50 &        1,367 \\
\hline
12 & In-context prompting \texttt{Qwen2.5-32B}          & 5.38\%  & 13.28\% & N/A   &  N/A    \\
13 & In-context prompting \texttt{Qwen2.5-32B-Instruct} & 9.62\%  & 13.28\% & N/A   &  N/A    \\
\hline
14 & \texttt{Qwen2.5-32B} on 50 examples                & 19.29\% & 33.59\% & 50    &  3,444  \\
\hline
15 & \texttt{Qwen2.5-Math-72B-Instruct}                 & 11.72\% &  16.14\% &  N/A & N/A     \\
\hline
\end{tabular}
\caption{Accuracy on Comp-Math-24-25 using non-reasoning and human-written data. Rows 1–5 report the SFT accuracy on non-reasoning data and its edited versions. Rows 6–11 present the SFT accuracy on human-written data and its edited versions. Rows 12–13 show the accuracy of few-shot prompting using \texttt{Qwen2.5-32B} and \texttt{Qwen2.5-32B-Instruct}. Row 14 reports the SFT accuracy of \texttt{Qwen2.5-32B} on 50 examples from \texttt{QwQ-32B-Preview}, while Row 15 shows the baseline accuracy of \texttt{Qwen2.5-Math-72B-Instruct}.}
\label{tab:eval-results-non-reasoning-data}
\end{table*}
\section{Human-Crafted Solutions with Structured Reasoning}

\subsection{Progressive Refinement of Human-Written Solutions}
\label{sec:human_written_data}
We created four progressively refined versions of solutions for the same 50 problems, with each version incorporating increasingly detailed and structured reasoning:
\begin{enumerate}
\item In the first version, we asked our volunteers, all experienced competitive math problem solvers, to provide comprehensive solutions to each problem. We emphasized the importance of explicitly documenting their thought process in great detail and instructed them not to imitate or copy the style of reasoning models such as DeepSeek R1. As a result, the reasoning in these solutions was articulated with significant depth, often surpassing the level of explanation typically found in textbooks or problem collections.
\item After reflecting on how the initial solutions could be improved to better elicit reasoning, we adopted a more structured format with clearly marked phases. Each solution was broken down into steps, using section headers such as “Let’s restate the problem”, “Let’s consider a straightforward approach”, “But wait…” (to indicate a mistake or obstacle), “Let’s try another approach” and finally “I think this approach works!” followed by “Final answer”. These labels helped signal the reasoning process explicitly and encouraged a more transparent, exploratory style.

\item For the third version, we drew inspiration from a teaching strategy often used with human students, solving significantly simplified versions of the problem or attempting to guess simple cases. We refer to this technique as “going back to the first grade.” In our experience, this strategy helps students better understand the underlying structure of the problem, often more effectively than merely restating it. By attempting to solve special cases or simplified versions, students (and models) can uncover key insights that are applicable to the general problem. To incorporate this idea, we added a new section immediately after “Let’s restate the problem,” beginning with: “Let's take some special cases”.

\item As we examined why reasoning model outputs often elicit stronger reasoning behavior than human-written solutions, we observed a key style difference. Model-generated solutions tend to follow an exploratory process, revealing intermediate steps as they emerge. For example, a model might find that $\angle A = \angle B$, then $\angle B = \angle C$, and finally conclude that $\triangle ABC$ is equilateral. In contrast, human solutions are usually composed after solving the problem through prior and informal reasoning. The final written solutions often reflect reconstructed reasoning, presenting conclusions before the supporting steps. This results in a non-causal narrative that masks the actual discovery process. To address this, we revised our solutions to enforce logical causality by presenting steps in the order they might be discovered. This restructuring aims to better mirror the reasoning style of models and more effectively prompt reasoning behavior in language models.
\end{enumerate}

Rows 6 to 10 in Table \ref{tab:eval-results-non-reasoning-data} show the SFT accuracy for the four versions of human-written data. Notably, the average length of these solutions is approximately 3K tokens, which is much longer than typical textbook solutions. However, even though we explicitly included reasoning steps and carefully documented human thought processes, the accuracy is still much lower than that of models trained on reasoning data. Even when compared to non-reasoning data, which lacks explicit reasoning, the human-written data achieves comparable \textbf{maj@64} scores but substantially lower \textbf{pass@1} scores.

We hypothesize that this is partly due to style variation in the human-written data. Since the solutions were authored by multiple individuals, there is considerable inconsistency in how reasoning is presented. In contrast, the non-reasoning and reasoning datasets generated by LLMs exhibit much more consistent and homogeneous styles, which likely help the model learn more stable reasoning patterns. The combination of limited data scale and high stylistic variance in the human-written data makes it more difficult for the model to generalize effectively. In contrast, the consistency in LLMs-generated data likely facilitates more effective learning, even under similar data size constraints.

\subsection{LLMs-Guided Editing of Human-Written Solutions}
We use the same first two prompts \ref{sec:data_edit_instruction} and \ref{sec:data_edit_few_shot} from Section~\ref{sec:solution_edit_non_reasoning_data}, excluding the third prompt since human-written data do not contain incorrect solutions, to revise the human-written data and encourage the insertion of reasoning behaviors.

Although we instructed the models to retain the original text and only insert reasoning behaviors, we still observed a decrease in the average length of the revised human-written solutions. We attribute this to the LLMs removing redundant or repetitive content, resulting in more concise and well-structured responses. As shown in Rows 10–11 of Table \ref{tab:eval-results-non-reasoning-data}, the SFT accuracy on the edited human-annotated data shows a slight improvement, ultimately aligning with the non-reasoning baseline but not surpassing it. We believe the improvements come from LLM-assisted editing, which helps make the structures more consistent, though not enough to to trigger reasoning behavior in the base model.

\section{Which Factors Influence Reasoning Ability in Training Data?}
\subsection{The Role of Solution Correctness in Reasoning Learning}
\label{sec:solution_correctness}
In the previous sections, we selected training data based on the correctness of the final answer, assuming that only complete and correct reasoning trajectories are beneficial for eliciting reasoning capabilities in base models. However, it remains unclear whether partially correct solutions that follow the correct output format but arrive at an incorrect final answer can still offer learning value. To explore this, we selected 50 solutions from the same data pool in Section \ref{sec:reasoning_data_generation} that produced the correct final answer format but were ultimately incorrect in their final conclusions. Despite the errors, it is possible many of these solutions contained valid intermediate reasoning steps that aligned with correct problem-solving strategies. 

The SFT results in Table \ref{tab:eval-results-on-reasoning-data} show that both the 50 correct (Row 8) and the 50 incorrect solutions (Row 1) were able to activate the base model’s reasoning ability to a comparable extent. This suggests that even solutions with incorrect final answers can provide valuable supervision signals, as they often contain structurally sound and logically coherent intermediate reasoning steps. Rather than simply memorizing knowledge from these examples, the model appears to benefit from exposure to the reasoning process itself, rather than what to think. These findings support the view that effective reasoning supervision does not necessarily require fully correct answers, as long as the underlying reasoning structure remains instructive.

\subsection{Does the Presence of Keywords Matter in Reasoning Data?}
Reasoning data often includes phrases such as “but wait”, “let me think” or “let me check” which are hypothesized to encourage behaviors like backtracking, self-verification, and reflection mechanisms believed to enhance a model’s reasoning ability. To assess the actual impact of these keywords, we took the same 50 solutions from \texttt{QwQ-32B-Preview} in Section \ref{sec:reasoning_data_generation}, and removed the top 10 most frequent reasoning-related phrases (e.g., “but wait”, “maybe I can consider” etc.).

The second row of Table~\ref{tab:eval-results-on-reasoning-data} shows that removing these keywords results in comparable performance, suggesting that the model’s reasoning ability is not primarily driven by the presence of such keywords. Instead, this supports the view that it is the underlying structure of high-quality reasoning demonstrations, rather than specific keywords, that activates the model’s reasoning capabilities. Notably, similar behavior has also been observed in~\cite{li2025llms}.

\subsection{Impact of Problem Difficulty on Model Reasoning}
\label{sec:problem_difficulty}
To investigate whether problem difficulty influences the utility of reasoning data, we used the pass rate of the 50 problems, calculated as the average success rate over 512 generations from the \texttt{QwQ-32B-Preview} model. Based on these scores, we categorized the problems into three difficulty levels: easy (pass ratio 0.7–1.0), medium (0.3–0.7), and hard (0.0–0.3). From each level, we randomly selected 10 problems and retained five correct solutions per problem, keeping the total dataset size consistent at 50. We matched the average solution length across difficulty levels, targeting approximately 3K tokens per solution. This process yielded three reasoning datasets (easy, medium, and hard), each containing 50 correct solutions with comparable lengths.

We fine-tuned the base model \texttt{Qwen2.5-32B} on each of these datasets. Interestingly, the resulting model performances (Rows 3 - 5 in Table \ref{tab:eval-results-on-reasoning-data}) were comparable across all difficulty levels, suggesting that problem difficulty alone does not significantly affect the model's ability to acquire reasoning skills. These findings are consistent with our previous observation: the presence of explicit reasoning demonstrations in the data is more critical than the difficulty of the problems themselves.
\begin{table*}[t]
\centering
\renewcommand{\arraystretch}{1.3} 
\begin{tabular}{rlcccc}
\hline
   & \textbf{Data} & \textbf{Pass@1} & \textbf{Maj@64} & \textbf{Average Length} \\
 \hline
 1  & Incorrect Solutions & 21.21\% & 33.59\%    & 3,303 \\
\hline
 2  & Without keywords & 18.70\% & 33.20\% &   3,279  \\
\hline
 3  & Difficulty Level (0.0 - 0.3)  & 19.38\% &35.94\% &   3,334  \\
 4  & Difficulty Level (0.3 - 0.7) & 20.95\% & 34.38\% &   3,822  \\
 5  & Difficulty Level (0.7 - 1) & 21.20\% & 31.64\% &    3,320 \\
\hline
 6  & 10 Problems & 21.19\% & 33.20\% &    3,445   \\
 7  & 25 Problems & 20.87\% & 33.59\% &    3,729   \\
 8  & 50 Problems & 19.29\% & 33.59\% &    3,433   \\
\hline
 9  & 2K average length & 19.67\%  &  26.95\%   &  2,421  \\
10  & 4K average length & 21.82\%  &  33.98\%   &  4,008  \\
11  & 6K average length & 23.30\%  &  34.77\%   &  6,049  \\
12  & 8K average length & 22.25\%  &  37.11\%   &  7,898  \\
\hline
\end{tabular}
\caption{Model accuracy on Comp-Math-24-25 under different reasoning dataset constructions, with each dataset containing 50 examples.}
\label{tab:eval-results-on-reasoning-data}
\end{table*}

\subsection{Impact of Problem Diversity on Model
Reasoning}
\label{sec:problem_difficulty}
To investigate the impact of problem diversity on the effectiveness of reasoning data, we fixed the total number of training examples at 50 while varying the number of unique problems. We still use the 50 problems set in Section~\ref{sec:problem_sets} and construct the dataset as follow: (1) 10 unique problems, each paired with 5 distinct solutions; (2) 25 unique problems, each with 2 distinct solutions; and (3) 50 unique problems, each with 1 solution. The 10 and 25 problem subsets were randomly sampled from the same pool of 50 problems. We reused the same generated solutions as in Section~\ref{sec:reasoning_data_generation}, filtering only for those with correct final answers.

Interestingly, we found that varying the number of unique problems, while keeping the total number of examples, had minimal impact on model performance (Rows 6 - 8 in Table \ref{tab:eval-results-on-reasoning-data}). Models trained on highly diverse datasets (50 problems × 1 solution) performed comparably to those trained on more repetitive ones (10 problems × 5 solutions). These results once again reinforce our earlier findings: it is the presence of high-quality reasoning demonstrations, rather than the diversity or difficulty of the problems, that most strongly drives reasoning behavior in models.

\subsection{The Influence of Solution Length on Reasoning Data Quality}
As we know, the most notable characteristic of reasoning data is its substantial length, which enables it to explore multiple reasoning paths and perform self-validation. To investigate the impact of example length on SFT performance, we constructed four datasets with lengths of 2K, 4K, 6K, and 8K tokens, using the same data pool described in Section~\ref{sec:reasoning_data_generation}. Since each problem has 512 generated solutions, we are able to filter correct solutions based on their length. Each dataset contains the same 50 problems, with one correct solution per problem. The only variation across datasets lies in the length of these solutions. The results in Table~\ref{tab:eval-results-on-reasoning-data} (Rows 9–12) show a slight performance improvement as the solution length increases. This improvement may be attributed to longer solutions exhibiting more frequent reasoning patterns and providing elaborate demonstrations of the reasoning process, thereby more effectively activating the model’s reasoning capabilities, especially given the limited dataset size.

\section{Related Work}
\subsection{Long CoT Reasoning}
Chain-of-Thought (CoT) prompting has been empirically shown to be an effective method for enhancing the reasoning capabilities of LLMs by encouraging the generation of intermediate logical steps. Initially introduced by Wei et al. \cite{wei2022chain}, this approach demonstrated strong performance across a wide range of reasoning tasks, including arithmetic computation, commonsense inference, and symbolic manipulation. While numerous follow-up studies (\cite{zhang2022automatic, wang2024chain, li2025structured, diao2023active, chia2023contrastive}) have sought to improve CoT-based reasoning in LLMs, many models still tend to generate reasoning traces that are short, shallow, or lacking in depth. This remains a key bottleneck in addressing complex tasks that demand more elaborate and iterative reasoning. Recently, researchers have introduced a variety of reasoning models, such as OpenAI o1 \cite{openai2024o1}, DeepSeek R1 \cite{guo2025deepseek}, and  Qwen QwQ \cite{qwq32b}, that are capable of generating long CoT solutions with strong self-validation and reflective reasoning patterns, enabling them to tackle complex reasoning tasks effectively. 

Two primary approaches have proven effective in eliciting LLMs to generate long CoT reasoning traces: knowledge distillation and reinforcement learning (RL). The former approach leverages advanced models such as DeepSeek R1 to produce long CoT traces, which are then used to supervise smaller models through SFT. This distillation-based strategy has resulted in a series of compact yet competitive reasoning models \cite{ye2025limo,moshkov2025aimo,muennighoff2025s1,novasky2025sky} that achieve performance comparable to DeepSeek R1. In the RL-based direction, reinforcement learning can be directly applied to base models or instruction-tuned models to enhance their reasoning capabilities. Several studies have successfully adopted this approach to reproduce reasoning models. For example, Yu et al. \cite{yu2025dapo} introduced an open-source RL algorithm that successfully transformed the Qwen2.5-32B model into a reasoning-oriented version. Similarly, Zhang et al. \cite{zhang2025srpo} proposed SRPO, a two-stage history-resampling policy optimization framework, which was also applied to Qwen2.5-32B and demonstrated superior performance over previous reasoning models with fewer training steps.

\subsection{Distillation for Math Reasoning}
Knowledge distillation from advanced models to smaller base models is an effective strategy for developing reasoning models that maintain strong performance while significantly reducing computational costs. DeepSeek \cite{guo2025deepseek} has released a series of distilled models that demonstrate exceptional performance in mathematical reasoning across various benchmarks, for example, DeepSeek-R1-Distill-Qwen-32B. Subsequent works have proposed to generate large-scale data from strong advanced models and then build mathematical reasoning models via distillation. Moshkov et al. \cite{moshkov2025aimo} constructed a large-scale dataset containing 3.2M long-form reasoning solutions based on 540,000 unique high-quality math problems. Their proposed model won the AI Mathematical Olympiad – Progress Prize 2. Similarly, Zhao et al. \cite{zhao20251} introduced a dataset of 1.4M question-response pairs accompanied by detailed reasoning traces. The models they developed outperformed the DeepSeek-R1-Distill-Llama-70B model across all math benchmarks. The Open R1 team \cite{openr12025math} also released a dataset of 220K examples, along with a SFT model distilled from DeepSeek-R1. In contrast, Muennighoff et al. \cite{muennighoff2025s1} developed the s1-32B model using a much smaller dataset of only 1K questions paired with reasoning traces, yet surpassed o1-preview on competition-level math problems by up to 27\%.

\section{Conclusion and Future Work}

In this work, we investigate how to induce reasoning behavior in a base language model based on different types of data. We first propose a simple and empirical definition of a reasoning model, based on the performance improvement of a base model after fine-tuning on a small number of high-quality reasoning traces, and demonstrate that as few as 20 reasoning examples can activate reasoning behavior in a base model. We then conduct extensive studies on two alternative data sources, including non-reasoning data and human-written solutions, and find that neither of them is sufficient to induce reasoning behavior in the base model, despite extensive prompt engineering and iterative refinement. Furthermore, we perform a series of ablation studies on the reasoning data and identify structural consistency in reasoning traces as a key factor in enabling the distillation of the reasoning ability.

For future work, one direction is to explore strategies that encourage deeper reasoning behaviors in LLMs, for example, providing partial solutions such as intermediate steps without final answers to prompt the model to complete the reasoning process on its own. We also plan to investigate other models and long CoT solutions to better understand model-specific reasoning behaviors. In addition, while this work focuses on mathematical problems, future studies could extend our framework to other reasoning-intensive domains such as symbolic logic, coding tasks, or scientific question answering, to examine the generalization of induced reasoning abilities. Finally, we aim to continuously refine human-written solutions by injecting more consistent reasoning patterns and harmonizing writing styles across annotators.

\bibliographystyle{plainnat}  
\bibliography{paper}  

\begin{thebibliography}{24}
\providecommand{\natexlab}[1]{#1}
\providecommand{\url}[1]{\texttt{#1}}
\expandafter\ifx\csname urlstyle\endcsname\relax
  \providecommand{\doi}[1]{doi: #1}\else
  \providecommand{\doi}{doi: \begingroup \urlstyle{rm}\Url}\fi

\bibitem[Chia et~al.(2023)Chia, Chen, Tuan, Poria, and Bing]{chia2023contrastive}
Yew~Ken Chia, Guizhen Chen, Luu~Anh Tuan, Soujanya Poria, and Lidong Bing.
\newblock Contrastive chain-of-thought prompting.
\newblock \emph{arXiv preprint arXiv:2311.09277}, 2023.

\bibitem[Diao et~al.(2023)Diao, Wang, Lin, Pan, Liu, and Zhang]{diao2023active}
Shizhe Diao, Pengcheng Wang, Yong Lin, Rui Pan, Xiang Liu, and Tong Zhang.
\newblock Active prompting with chain-of-thought for large language models.
\newblock \emph{arXiv preprint arXiv:2302.12246}, 2023.

\bibitem[Guo et~al.(2025)Guo, Yang, Zhang, Song, Zhang, Xu, Zhu, Ma, Wang, Bi, et~al.]{guo2025deepseek}
Daya Guo, Dejian Yang, Haowei Zhang, Junxiao Song, Ruoyu Zhang, Runxin Xu, Qihao Zhu, Shirong Ma, Peiyi Wang, Xiao Bi, et~al.
\newblock {DeepSeek-R1: Incentivizing Reasoning Capability in LLMs via Reinforcement Learning}.
\newblock \emph{arXiv preprint arXiv:2501.12948}, 2025.

\bibitem[Li et~al.(2025{\natexlab{a}})Li, Cao, Griggs, Liu, Mo, Tang, Hegde, Hakhamaneshi, Patil, Zaharia, et~al.]{li2025llms}
Dacheng Li, Shiyi Cao, Tyler Griggs, Shu Liu, Xiangxi Mo, Eric Tang, Sumanth Hegde, Kourosh Hakhamaneshi, Shishir~G Patil, Matei Zaharia, et~al.
\newblock Llms can easily learn to reason from demonstrations structure, not content, is what matters!
\newblock \emph{arXiv preprint arXiv:2502.07374}, 2025{\natexlab{a}}.

\bibitem[Li et~al.(2025{\natexlab{b}})Li, Li, Li, and Jin]{li2025structured}
Jia Li, Ge~Li, Yongmin Li, and Zhi Jin.
\newblock Structured chain-of-thought prompting for code generation.
\newblock \emph{ACM Transactions on Software Engineering and Methodology}, 34\penalty0 (2):\penalty0 1--23, 2025{\natexlab{b}}.

\bibitem[Loshchilov and Hutter(2019)]{loshchilov2017decoupled}
Ilya Loshchilov and Frank Hutter.
\newblock {Decoupled Weight Decay Regularization}.
\newblock In \emph{ICLR}, 2019.

\bibitem[Moshkov et~al.(2025)Moshkov, Hanley, Sorokin, Toshniwal, Henkel, Schifferer, Du, and Gitman]{moshkov2025aimo}
Ivan Moshkov, Darragh Hanley, Ivan Sorokin, Shubham Toshniwal, Christof Henkel, Benedikt Schifferer, Wei Du, and Igor Gitman.
\newblock Aimo-2 winning solution: Building state-of-the-art mathematical reasoning models with openmathreasoning dataset.
\newblock \emph{arXiv preprint arXiv:2504.16891}, 2025.

\bibitem[Muennighoff et~al.(2025)Muennighoff, Yang, Shi, Li, Fei-Fei, Hajishirzi, Zettlemoyer, Liang, Cand{\`e}s, and Hashimoto]{muennighoff2025s1}
Niklas Muennighoff, Zitong Yang, Weijia Shi, Xiang~Lisa Li, Li~Fei-Fei, Hannaneh Hajishirzi, Luke Zettlemoyer, Percy Liang, Emmanuel Cand{\`e}s, and Tatsunori Hashimoto.
\newblock s1: Simple test-time scaling.
\newblock \emph{arXiv preprint arXiv:2501.19393}, 2025.

\bibitem[of~America()]{AMC_MAA}
Mathematical~Association of~America.
\newblock {American Mathematics Competitions}.
\newblock \url{https://maa.org/student-programs/amc/}.

\bibitem[of~Problem~Solving()]{AoPS_Forums}
Art of~Problem~Solving.
\newblock {AoPS Forums}.
\newblock \url{https://artofproblemsolving.com/community/}.

\bibitem[OpenAI(2024)]{openai2024o1}
OpenAI.
\newblock Openai o1 preview.
\newblock \url{https://openai.com/o1/}, 2024.

\bibitem[Shen et~al.(2024)Shen, Wang, Delalleau, Zeng, Dong, Egert, Sun, Zhang, Jain, Taghibakhshi, et~al.]{shen2024nemo}
Gerald Shen, Zhilin Wang, Olivier Delalleau, Jiaqi Zeng, Yi~Dong, Daniel Egert, Shengyang Sun, Jimmy Zhang, Sahil Jain, Ali Taghibakhshi, et~al.
\newblock Nemo-aligner: Scalable toolkit for efficient model alignment.
\newblock \emph{arXiv preprint arXiv:2405.01481}, 2024.

\bibitem[Team(2025{\natexlab{a}})]{novasky2025sky}
NovaSky Team.
\newblock Sky-t1: Fully open-source reasoning model with o1-preview performance in \$450 budget.
\newblock \url{https://novasky-ai.github.io/posts/sky-t1}, 2025{\natexlab{a}}.
\newblock Accessed: 2025-01-09.

\bibitem[Team(2025{\natexlab{b}})]{openr12025math}
OpenR1 Team.
\newblock Open r1 math 200k.
\newblock \url{https://huggingface.co/datasets/open-r1/OpenR1-Math220k}, February 2025{\natexlab{b}}.
\newblock Accessed: 2025-05-19.

\bibitem[Team(2025{\natexlab{c}})]{qwq32b}
Qwen Team.
\newblock {QwQ-32B: Embracing the Power of Reinforcement Learning}, March 2025{\natexlab{c}}.
\newblock URL \url{https://qwenlm.github.io/blog/qwq-32b/}.

\bibitem[Wang and Zhou(2024)]{wang2024chain}
Xuezhi Wang and Denny Zhou.
\newblock Chain-of-thought reasoning without prompting.
\newblock \emph{arXiv preprint arXiv:2402.10200}, 2024.

\bibitem[Wei et~al.(2022)Wei, Wang, Schuurmans, Bosma, Xia, Chi, Le, Zhou, et~al.]{wei2022chain}
Jason Wei, Xuezhi Wang, Dale Schuurmans, Maarten Bosma, Fei Xia, Ed~Chi, Quoc~V Le, Denny Zhou, et~al.
\newblock Chain-of-thought prompting elicits reasoning in large language models.
\newblock \emph{NeurIPS}, 2022.

\bibitem[Yang et~al.(2024)Yang, Zhang, Hui, Gao, Yu, Li, Liu, Tu, Zhou, Lin, Lu, Xue, Lin, Liu, Ren, and Zhang]{yang2024qwen25mathtechnicalreportmathematical}
An~Yang, Beichen Zhang, Binyuan Hui, Bofei Gao, Bowen Yu, Chengpeng Li, Dayiheng Liu, Jianhong Tu, Jingren Zhou, Junyang Lin, Keming Lu, Mingfeng Xue, Runji Lin, Tianyu Liu, Xingzhang Ren, and Zhenru Zhang.
\newblock {Qwen2.5-Math Technical Report: Toward Mathematical Expert Model via Self-Improvement}, 2024.

\bibitem[Yang et~al.(2025)Yang, Yang, Zhang, Hui, Zheng, Yu, Li, Liu, Huang, Wei, Lin, Yang, Tu, Zhang, Yang, Yang, Zhou, Lin, Dang, Lu, Bao, Yang, Yu, Li, Xue, Zhang, Zhu, Men, Lin, Li, Tang, Xia, Ren, Ren, Fan, Su, Zhang, Wan, Liu, Cui, Zhang, and Qiu]{qwen2025qwen25technicalreport}
An~Yang, Baosong Yang, Beichen Zhang, Binyuan Hui, Bo~Zheng, Bowen Yu, Chengyuan Li, Dayiheng Liu, Fei Huang, Haoran Wei, Huan Lin, Jian Yang, Jianhong Tu, Jianwei Zhang, Jianxin Yang, Jiaxi Yang, Jingren Zhou, Junyang Lin, Kai Dang, Keming Lu, Keqin Bao, Kexin Yang, Le~Yu, Mei Li, Mingfeng Xue, Pei Zhang, Qin Zhu, Rui Men, Runji Lin, Tianhao Li, Tianyi Tang, Tingyu Xia, Xingzhang Ren, Xuancheng Ren, Yang Fan, Yang Su, Yichang Zhang, Yu~Wan, Yuqiong Liu, Zeyu Cui, Zhenru Zhang, and Zihan Qiu.
\newblock {Qwen2.5 Technical Report}, 2025.
\newblock URL \url{https://arxiv.org/abs/2412.15115}.

\bibitem[Ye et~al.(2025)Ye, Huang, Xiao, Chern, Xia, and Liu]{ye2025limo}
Yixin Ye, Zhen Huang, Yang Xiao, Ethan Chern, Shijie Xia, and Pengfei Liu.
\newblock Limo: Less is more for reasoning.
\newblock \emph{arXiv preprint arXiv:2502.03387}, 2025.

\bibitem[Yu et~al.(2025)Yu, Zhang, Zhu, Yuan, Zuo, Yue, Fan, Liu, Liu, Liu, et~al.]{yu2025dapo}
Qiying Yu, Zheng Zhang, Ruofei Zhu, Yufeng Yuan, Xiaochen Zuo, Yu~Yue, Tiantian Fan, Gaohong Liu, Lingjun Liu, Xin Liu, et~al.
\newblock Dapo: An open-source llm reinforcement learning system at scale.
\newblock \emph{arXiv preprint arXiv:2503.14476}, 2025.

\bibitem[Zhang et~al.(2025)Zhang, Wang, Cheng, Zhuang, Lin, Zhang, Wang, Cui, Wang, Peng, et~al.]{zhang2025srpo}
Xiaojiang Zhang, Jinghui Wang, Zifei Cheng, Wenhao Zhuang, Zheng Lin, Minglei Zhang, Shaojie Wang, Yinghan Cui, Chao Wang, Junyi Peng, et~al.
\newblock Srpo: A cross-domain implementation of large-scale reinforcement learning on llm.
\newblock \emph{arXiv preprint arXiv:2504.14286}, 2025.

\bibitem[Zhang et~al.(2022)Zhang, Zhang, Li, and Smola]{zhang2022automatic}
Zhuosheng Zhang, Aston Zhang, Mu~Li, and Alex Smola.
\newblock Automatic chain of thought prompting in large language models.
\newblock \emph{arXiv preprint arXiv:2210.03493}, 2022.

\bibitem[Zhao et~al.(2025)Zhao, Wang, Peng, Zhao, Tian, Chen, Ji, and Li]{zhao20251}
Han Zhao, Haotian Wang, Yiping Peng, Sitong Zhao, Xiaoyu Tian, Shuaiting Chen, Yunjie Ji, and Xiangang Li.
\newblock {1.4 Million Open-Source Distilled Reasoning Dataset to Empower Large Language Model Training}.
\newblock \emph{arXiv preprint arXiv:2503.19633}, 2025.

\end{thebibliography}

\appendix

\onecolumn
\section{Data Generation}
\label{sec:LLM_prompts}

\subsection{Data Generation with Instruction}
\label{sec:data_generation_instruction}

\begin{tcolorbox}[breakable,width=\textwidth,colback=white,colframe=NvidiaGreen,title={\centering \large  \textbf{Prompt: Data Generation with Instruction}}]
\footnotesize                  
\begin{lstlisting}[
breaklines=true, postbreak={},breakindent=0pt, 
label={lst:math-prompt}]
I will give you a math problem and ask to solve it. Your goal is to write your thinking process as you approach this math problem. 
  Solve the following mathematical problem in a natural, exploratory, and human-like reasoning style. 
  Your reasoning should reflect how a person would think through the problem step by step, rather than providing a rigid, structured answer. 
  Follow these guidelines:
  
  1. Think aloud and reason through the problem dynamically:
     Start with intuitive observations and build from there.
     Use a conversational tone, as if you are thinking through the problem in real time.
     Include phrases like:
        "Hmm, let me see..."
        "Wait, does this actually work?"
        "Oh, I think I see a pattern!"
  
  2. Incorporate self-doubt and self-correction:
     If you realize a mistake or reconsider an approach, explicitly acknowledge it.
     Say things like:
        "Wait, no, that can't be right. Let me check again."
        "Actually, I think I missed something there."
        "Hold on, let me backtrack and see where I went wrong."
  

  3. Use trial and error before arriving at a final conclusion:
    Try different approaches and compare them.
    If an assumption is made, question whether it's valid.
    Example phrases:
       "Let's try assuming x=y=z and see what happens."
       "Hmm, this simplifies things, but does it actually lead to a solution?"
  
  4. Verify and reflect on the final answer:
     Once you reach an answer, check if it actually satisfies the conditions.
     Explicitly validate by plugging values back in.
     Include phrases like:
        "Alright, let's plug this back in and see if it holds up."
        "Yep, this checks out seems like we got the right answer"
        "Wait, let's test it with a smaller case like n=2 to make sure."
  
  5. Make the reasoning process engaging and organic:
      Avoid overly rigid or textbook-style explanations.
      Feel free to pause, rethink, and approach things differently.
      Keep the explanation engaging, as if you're discussing it with a friend.
  

  Finish the solution with **Final Answer**
  \[ \boxed{{YOUR ANSWER}} \] 

  {problem}


  Please share your thinking process now. Remember: your goal is to write your thinking process as you approach this math problem. Note that we are not that much interested in the final solution, but much more in the process that lead you to it. 
  Did you have references to other problems you solved before in your mind? Write that down. Did you ponder upon which approach to take before doing derivations? Write that down. Does your current approach look too complicated and you need to try something else? 
  Say that and write down any alternative ideas you have. Did you notice that something doesn't look good and you need to double check previous derivations for mistake? That's what we are looking for - write that down and correct the solution, but don't erase the mistake! 
  The more detailed this is, the better.
\end{lstlisting}
\end{tcolorbox}

\subsection{Data Generation with Few-Shot Instruction}
\label{sec:data_generation_few_shot}

\begin{tcolorbox}[breakable,width=\textwidth,colback=white,colframe=NvidiaGreen,title={\centering \large  \textbf{Prompt: Data Generation with Few-Shot Instruction}}]
\footnotesize                  
\begin{lstlisting}[
breaklines=true, postbreak={},breakindent=0pt, 
label={lst:math-prompt-invalid}]
I will give you a math problem and ask to solve it. Your goal is to write your thinking process as you approach this math problem. 
  Solve the following mathematical problem in a natural, exploratory, and human-like reasoning style. 
  Your reasoning should reflect how a person would think through the problem step by step, rather than providing a rigid, structured answer. 
  Follow these guidelines:
  
  1. Think aloud and reason through the problem dynamically:
     Start with intuitive observations and build from there.
     Use a conversational tone, as if you are thinking through the problem in real time.
     Include phrases like:
        "Hmm, let me see..."
        "Wait, does this actually work?"
        "Oh, I think I see a pattern!"
  
  2. Incorporate self-doubt and self-correction:
     If you realize a mistake or reconsider an approach, explicitly acknowledge it.
     Say things like:
        "Wait, no, that can't be right. Let me check again."
        "Actually, I think I missed something there."
        "Hold on, let me backtrack and see where I went wrong."
  

  3. Use trial and error before arriving at a final conclusion:
    Try different approaches and compare them.
    If an assumption is made, question whether it's valid.
    Example phrases:
       "Let's try assuming x=y=z and see what happens."
       "Hmm, this simplifies things, but does it actually lead to a solution?"
  
  4. Verify and reflect on the final answer:
     Once you reach an answer, check if it actually satisfies the conditions.
     Explicitly validate by plugging values back in.
     Include phrases like:
        "Alright, let's plug this back in and see if it holds up."
        "Yep, this checks out, seems like we got the right answer!"
        "Wait, let's test it with a smaller case like n=2 to make sure."
  
  5. Make the reasoning process engaging and organic:
      Avoid overly rigid or textbook-style explanations.
      Feel free to pause, rethink, and approach things differently.
      Keep the explanation engaging, as if you're discussing it with a friend.
  

  Finish the solution with **Final Answer**
  \[ \boxed{{YOUR ANSWER}} \] 



  I first provide you with the following examples of how to solve math problems and then provide you a problem you need to solve.

  {examples}


  This is the problem you need to solve:
  {problem}


  Please share your thinking process now. Remember: your goal is to write your thinking process as you approach this math problem. Note that we are not that much interested in the final solution, but much more in the process that lead you to it. 
  Did you have references to other problems you solved before in your mind? Write that down. Did you ponder upon which approach to take before doing derivations? Write that down. Does your current approach look too complicated and you need to try something else? 
  Say that and write down any alternative ideas you have. Did you notice that something doesn't look good and you need to double check previous derivations for mistake? That's what we are looking for - write that down and correct the solution, but don't erase the mistake! 
  The more detailed this is, the better.
\end{lstlisting}
\end{tcolorbox}

\section{Data Edition}
\subsection{Data Edit with Instruction}
\label{sec:data_edit_instruction}

\begin{tcolorbox}[breakable,width=\textwidth,colback=white,colframe=NvidiaGreen,title={\centering \large  \textbf{Prompt: Data Edition with Instruction}}]
\footnotesize                  
\begin{lstlisting}[
breaklines=true, postbreak={},breakindent=0pt, 
label={lst:math-prompt-mcq}]
You will be provided with a mathematical solution.
Your task is to revise it by weaving in self-questioning, step-by-step verification, and natural self-correction.
Keep the original content and structure intact, but enhance the reasoning process with thoughtful self-reflection.

Instructions:

1. Preserve the format and logical flow of the original solution.
    Do not add extra sections or headers that might interfere with smooth integration.

2. Incorporate at least 3 - 5 self verification moments throughout the reasoning. Use natural phrasing such as:

    Wait, something seems off here, let me double check.
    
    Hold on, did I consider all the possible cases?
    
    Let's go back and verify this step before moving on.

3. Stay aligned with the original solution, ensuring a seamless blend with the surrounding content.



Now, I will provide you with a solution. Please enhance it by incorporating self-questioning, verification, and natural self-correction to ensure both rigor and clarity.

Solution: {solution}



  
\end{lstlisting}
\end{tcolorbox}

\subsection{Data Edit with Few-Shot Instruction}
\label{sec:data_edit_few_shot}

\begin{tcolorbox}[breakable,width=\textwidth,colback=white,colframe=NvidiaGreen,title={\centering \large  \textbf{Prompt: Data Edit with Few-Shot Instruction}}]
\footnotesize                  
\begin{lstlisting}[
breaklines=true, postbreak={},breakindent=0pt, 
label={lst:math-prompt-mcq}]
You will be provided with a mathematical solution.
   Your task is to revise it by weaving in self-questioning, step-by-step verification, and natural self-correction.
   Keep the original content and structure intact, but enhance the reasoning process with thoughtful self-reflection.

   Instructions:

   1. Preserve the format and logical flow of the original solution.
        Do not add extra sections or headers that might interfere with smooth integration.

   2. Incorporate at least 35 self verification moments throughout the reasoning. Use natural phrasing such as:

        Wait, something seems off here, let me double check.
    
    Hold on, did I consider all the possible cases?
    
    Let's go back and verify this step before moving on.

    3. Stay aligned with the original solution, ensuring a seamless blend with the surrounding content.


    I will provide you with a few examples. 
    Please refer to these examples as template and follow the instructions above when revising the solution I give you afterward.

  
  {examples}
  
  Now, I will provide you with a solution. Please enhance it by incorporating self-questioning, verification, and natural self-correction to ensure both rigor and clarity.

  Solution: {solution}



  
\end{lstlisting}
\end{tcolorbox}

\subsection{Data Edit with Incorrect Solution Combination}
\label{sec:data_edit_combination}

\begin{tcolorbox}[breakable,width=\textwidth,colback=white,colframe=NvidiaGreen,title={\centering \large  \textbf{Prompt: Data Edit with Incorrect Solution Combination}}]
\footnotesize                  
\begin{lstlisting}[
breaklines=true, postbreak={},breakindent=0pt, 
label={lst:math-prompt-mcq}]
You are an advanced mathematical reasoning assistant.
Your goal is to merge a correct solution while incorporating elements from an incorrect solution to create a cohesive, logically structured response that includes natural self-corrections.
The final answer must remain correct while the reasoning process should be enhanced through logical self-reflection and error correction.
please start with your modified output with "Solution"
please make sure finish the solution with **Final Answer**
\[ \boxed{{YOUR ANSWER}} \] 

Instructions:
1. Analyze the multiple correct solutions and consolidate them into a single refined version that preserves all key insights.

2. Identify mistakes from the incorrect solutions (e.g., calculation errors, conceptual misunderstandings, missing steps).

3. Inject 3 - 6 subtle mistakes from the incorrect solutions into the reasoning process.

4. Ensure these mistakes are discovered and corrected naturally using logical problem-solving (not explicitly stated as intentional errors).

5. Use a natural and engaging style for self-correction, incorporating phrases like:
Wait, something feels off here... let me check.
"Oh! I think I see the mistake...
"Hold on, I might have miscalculated this part."
Let's go back and check this, something doesn't look right.

6. Maintain clarity, logical flow, and rigor in the final solution.

7. Ensure the final answer is correct, even after introducing and correcting the errors.



Please avoid using phrases like "let us introduce a few mistakes," "we will intentionally add some errors," or "for demonstration purposes, 
we will insert some mistakes," and ensure the wording sounds natural.

Now, I will provide one correct solution and one incorrect solution. Please merge them into a single solution following the instructions above. 
Ensure that the final response integrates insights from the correct solutions while incorporating and naturally correcting errors from the incorrect solution through logical self-correction


Correct solutions:

Correct solution : {correct} 
Incorrect solution : {incorrect} 
\end{lstlisting}
\end{tcolorbox}

\section{SFT Accuracy for Qwen2.5-7B and Qwen2.5-14B}
\label{sec:sft_7b_14b}
\begin{figure}[htbp]
    \centering
        \includegraphics[width=0.5\textwidth]{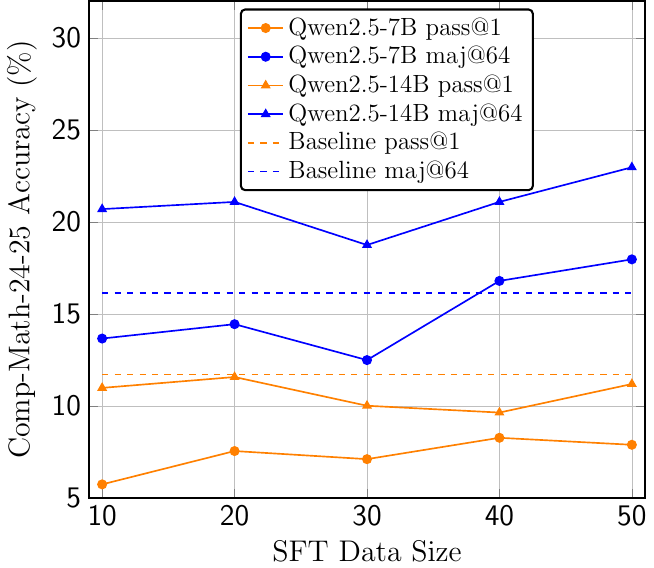}
    \caption{SFT accuracy of \texttt{Qwen2.5-7B} and \texttt{Qwen2.5-14B} on Comp-Math-24-25 under varying data sizes.}
    \label{fig:performance_varying_number_7b_14b}
\end{figure}
We use the same data as in Section~\ref{sec:reasoning_data_generation} and conduct SFT on \texttt{Qwen2.5-7B} and \texttt{Qwen2.5-14B}, using the same training and evaluation settings as \texttt{Qwen2.5-32B}. The SFT accuracy results are shown in Figure~\ref{fig:performance_varying_number_7b_14b}. The results show that \texttt{Qwen2.5-7B} performs below the baseline, while \texttt{Qwen2.5-14B} only slightly outperforms it. However, unlike \texttt{Qwen2.5-32B}, neither model is able to effectively enter reasoning mode with only minimal supervision and a limited number of high-quality, long CoT examples. 

\section{SFT Accuracy under Hyperparameters}
\label{sec:grid_search}
We performed a grid search for hyperparameter tuning on \texttt{Qwen2.5-32B} using 50 reasoning examples, with the results summarized in Table \ref{tab:hyperpara_performance}. Since we found that a learning rate of 1e-5 and a step size of 50 yielded the best performance, we fixed these hyperparameters for all experiments presented in this paper.

\begin{table}[h!]
\centering
\renewcommand{\arraystretch}{1.5} 
\begin{tabular}{lcc}
\hline
\textbf{Hyperparameters} & \textbf{pass@1}  & \textbf{maj@64} \\
\hline
LR = 1e-4, Steps = 50 & 6.52\% & 12.50\% \\

\hline
LR = 1e-5, Steps = 50  & 19.29\% & 33.59\% \\
\hline
LR = 1e-6, Steps = 50  & 8.53\% & 18.36\% \\
\hline
LR = 1e-5, Steps = 100 & 18.60\% & 31.64\% \\
\hline
LR = 1e-5, Steps = 200 & 15.05\% & 32.42\% \\
\hline
\end{tabular}
\caption{SFT accuracy on Comp-Math-24-25 under varying hyperparameters.}
\label{tab:hyperpara_performance}
\end{table}

\end{document}